\documentclass{article}

\usepackage{caption}
\usepackage{PRIMEarxiv}
\usepackage[utf8]{inputenc} 
\usepackage[T1]{fontenc}    
\usepackage{hyperref}       
\usepackage{url}            
\usepackage{booktabs}       
\usepackage{amsfonts}       
\usepackage{nicefrac}       
\usepackage{microtype}      
\usepackage{lipsum}
\usepackage{fancyhdr}       
\usepackage{graphicx}       
\usepackage{amsmath}
\usepackage{xcolor}
\graphicspath{{media/}}     
\usepackage{multirow}
\usepackage{cite}
\usepackage{footmisc}

\usepackage{CJKutf8}

\title{Exploring OCR Capabilities of GPT-4V(ision) : \\ A Quantitative and In-depth Evaluation 
}

\author{
  \parbox{\linewidth}{\centering
Yongxin Shi, Dezhi Peng, Wenhui Liao, Zening Lin, Xinhong Chen, Chongyu Liu, Yuyi Zhang, Lianwen Jin\thanks{Corresponding Author.}
  }
  \\
  \parbox{\linewidth}{\centering\vspace{2mm}
  South China University of Technology~~\\
  }
 \\
 \parbox{\linewidth}{\centering\vspace{2mm}
 \tt yongxin$\_$shi@foxmail.com, eelwjin@scut.edu.cn}
 \\
}

\begin{document}
\maketitle

\begin{abstract}

This paper presents a comprehensive evaluation of the Optical Character Recognition (OCR) capabilities of the recently released GPT-4V(ision), a Large Multimodal Model (LMM). We assess the model's performance across a range of OCR tasks, including scene text recognition, handwritten text recognition, handwritten mathematical expression recognition, table structure recognition, and information extraction from visually-rich document. The evaluation reveals that GPT-4V performs well in recognizing and understanding Latin contents, but struggles with multilingual scenarios and complex tasks. Specifically, it showed limitations when dealing with non-Latin languages and complex tasks such as handwriting mathematical expression recognition, table structure recognition, and end-to-end semantic entity recognition and pair extraction from document image. Based on these observations, we affirm the necessity and continued research value of specialized OCR models. In general, despite its versatility in handling diverse OCR tasks, GPT-4V does not outperform existing state-of-the-art OCR models. How to fully utilize pre-trained general-purpose LMMs such as GPT-4V for OCR downstream tasks remains an open problem.
The study offers a critical reference for future research in OCR with LMMs.
Evaluation pipeline and results are available at~\url{https://github.com/SCUT-DLVCLab/GPT-4V_OCR}.


\end{abstract}


\section{Introduction}
The emergence of ChatGPT~\cite{chatgpt} marks a significant milestone in the field of Artificial Intelligence (AI). 
Concurrently, it has ignited a surge in Large Language Models (LLMs) research across both academia and industry, with models such as GLM-130B~\cite{zeng2022glm}, Alpaca~\cite{alpaca}, Vicuna~\cite{chiang2023vicuna}, LLaMA~\cite{touvron2023llama}, ERNIE Bot~\cite{wenxin}, Qwen~\cite{tongyi}, Baichuan2~\cite{baichuan2023baichuan2}.
The success of LLMs has also spurred the development of Large Multimodal Models (LMMs).
Many initiatives are now striving to expand the multimodal capabilities of LLMs, including BLIP-2~\cite{li2023blip2}, OpenFlamingo~\cite{anas_awadalla_2023_7733589}, LLaVA~\cite{liu2023visual}, MiniGPT4~\cite{zhu2023minigpt4}, and mPLUG-Owl~\cite{ye2023mplugowl}.

Particularly, the recent release of GPT-4V(ision)~\cite{yang2023dawn} presents a significant breakthrough in the domain of LMMs. 
Researchers across diverse fields are eager to comprehend the capabilities of GPT-4V, with those in the Optical Character Recognition (OCR) domain displaying particular curiosity in its potential to address OCR tasks.
While the official report qualitatively demonstrates GPT-4V's abilities in several OCR-related tasks (including text recognition, expression recognition, and document understanding), quantitative assessment and in-depth analysis are urgently needed, which will provide valuable insights and essential references for future research.

To this end, we conduct a quantitative evaluation of GPT-4V on mainstream OCR tasks, including Scene Text Recognition (STR)~\cite{shi2016end,shi2018aster,luo2019moran,wang2020decoupled, liao2020mask,fang2021read,liu2021abcnet,zhong2022sgbanet,huang2022swintextspotter,ye2023deepsolo,huang2023estextspotter,lyu2018mask,liao2019mask,peng2022spts,liu2023sptsv2,zhang2023arbitrary}, Handwritten Text Recognition (HTR)~\cite{wang2011handwritten, bluche2016joint, xie2017learning, peng2019fast, yousef2020origaminet, peng2022recognition, peng2022pagenet, huang2023segctc, coquenet2023dan}, Handwritten Mathematical Expression Recognition (HMER)~\cite{zhang2017watch, zhang2018multi, yuan2022syntax, zhong2022tree, bian2022handwritten, zhao2022comer, li2022counting, li2023improving}, Table Structure Recognition (TSR)~\cite{zhong2020image, ye2021pingan, liu2021show, liu2022neural, chen2022complex, lin2022tsrformer, huang2023improving, shen2023divide}, and Information Extraction from Visually-rich Document (VIE)~\cite{9412927, wang-etal-2022-lilt, Wang_Liu_Jin_Tang_Zhang_Zhang_Wang_Wu_Cai_2021, DBLP:conf/ijcai/WangWTJMDH21, 10.1145/3394486.3403172, 10.1145/3503161.3548112, DBLP:conf/aaai/HongKJHNP22, Li_2021_CVPR, Appalaraju_2021_ICCV, yu2023structextv}. For the above tasks, we employ some commonly used benchmarks in the OCR domain for evaluation: \textbf{(1) STR:} CUTE80~\cite{Risnumawan2014ARA}, SCUT-CTW1500~\cite{Liu2019CurvedST}, Total-Text~\cite{Chng2017TotalTextAC}, WordArt~\cite{xie2022toward}, ReCTS~\cite{zhang2019icdar} and MLT19~\cite{nayef2019icdar2019}, 
\textbf{(2) HTR:} IAM~\cite{marti2002iam} and CASIA-HWDB~\cite{yin2013icdar},
\textbf{(3) HMER:} CROHME2014~\cite{mouchere2014icfhr} and HME100K~\cite{yuan2022syntax}, 
\textbf{(4) TSR:} SciTSR~\cite{chi2019complicated} and WTW~\cite{long2021parsing}, 
\textbf{(5) VIE:} FUNSD~\cite{jaume2019funsd} and XFUND~\cite{xu2022xfund} Chinese subset (XFUND-zh).

The evaluation results suggest that GPT-4V does not match the performance of specialized OCR models. Specifically, GPT-4V demonstrates superior performance in Latin content but encounters limitations when dealing with other languages.
Furthermore, GPT-4V struggles in complex scenarios for tasks such as HMER, TSR, and VIE.

Based on the experimental results, we try to address an important question: 
\textbf{do specialized models still hold research value in the OCR field?} 
Given the three critical drawbacks of GPT-4V, namely, limited performance in multilingual and complex scenarios, high inference costs, and challenges in updating, 
we argue that existing LMMs struggle to simultaneously handle various OCR tasks~\cite{liu2023hidden}. Therefore, we affirm the \textit{\textbf{continued research value}} of specialized models in the OCR field.
However, it is still crucial to leverage the potential of LMMs like GPT-4V for future OCR research. There may be three potential directions worth investigating, including semantic understanding enhancement, downstream task finetuning, and auto/semi-auto data construction.




\section{Experiments}
\label{Experiments}
We evaluate GPT-4V on the following OCR tasks: scene text recognition, handwritten text recognition, handwritten mathematical expression recognition, table structure recognition, and information extraction from visually-rich document.
The evaluation process was conducted within the web-based dialogue interface with GPT-4V, of which we directly uploaded the image and prompt, and then extracted relevant answers from the generated responses. 
The prompts for each task were meticulously designed.
Additionally, to prevent interference from contextual information, we used a separate dialogue window for each image. 
Due to the conversation limits (50 conversations per 3 hours) of GPT-4V, we conducted sampling on datasets with a large number of samples.

\subsection{Scene text recognition} 
\label{Sec:STR}
\paragraph{Dataset} We focus on both word-level text recognition and end-to-end text spotting. 
For word-level text recognition, we employ CUTE80~\cite{Risnumawan2014ARA}, SCUT-CTW1500~\cite{Liu2019CurvedST}, Total-Text~\cite{Chng2017TotalTextAC}, WordArt~\cite{xie2022toward} in English and ReCTS~\cite{zhang2019icdar} in Chinese. We randomly select 50 images from each dataset above for evaluation. The datasets are downloaded from \footnote{\url{https://github.com/Yuliang-Liu/MultimodalOCR}}.
 
\begin{itemize}
    \item \textbf{CUTE80} comprises 80 images specifically curated for the purpose of evaluating curved text.
    \item \textbf{SCUT-CTW1500} is a comprehensive curved text dataset encompassing a total of 1500 images.
    \item \textbf{Total-Text} has 1,555 scene images which collected with curved text in mind.
    \item \textbf {WordArt} consists of 6316 artistic text images, which primarily features challenging artistic text.
    \item \textbf{ReCTS} is a large-scale dataset of 25,000 images, which mainly focuses on reading Chinese text on signboard.
\end{itemize}

In the end-to-end text spotting task, we use MLT19~\cite{nayef2019icdar2019} to evaluate the multilingual capabilities of GPT-4V. 
For each language, we randomly select 20 images from the training set. 
Additionally, to investigate the impact of image resolution on recognition results, we select 20 English images from the aforementioned subset and resize their long sides to 128, 256, 512, 1024, and 2048 pixels, respectively.
\begin{itemize}
    \item \textbf{MLT19} is a dataset for Multi-Lingual scene Text (MLT) detection and recognition, which consists of 20,000 images containing text from 10 languages.
\end{itemize}

\paragraph{Prompt} For word-level English text recognition, we use the following prompt:
\textit{"What is the scene text in the image?"}, while for ReCTS in Chinese, we translate the prompt into Chinese, resulting in:
\begin{CJK*}{UTF8}{gbsn}
"图片中的场景文字是什么？"
\end{CJK*}
The prompt in end-to-end text spotting is:
\textit{"What are all the scene text in the image? Do not translate."}

\paragraph{Metric} For the evaluation of word-level recognition, we employ word accuracy ignoring case and symbols (WAICS)~\cite{jiang2023revisiting} as metric. 
In the task of end-to-end text spotting, the predictions of GPT-4V and ground truths (GT) are split with spaces and then evaluated using precision and recall. Precision represents the ratio of correctly identified words to those generated by GPT-4V, while recall is the ratio of correctly identified words to the total number of GT words. We also compute the F1 score as follow.

\begin{equation}
F1 = \frac{2 \cdot \text{precision} \cdot \text{recall}}{\text{precision} + \text{recall}}
\end{equation}

\begin{figure*}[h]
\centering
\includegraphics[width=1\textwidth]{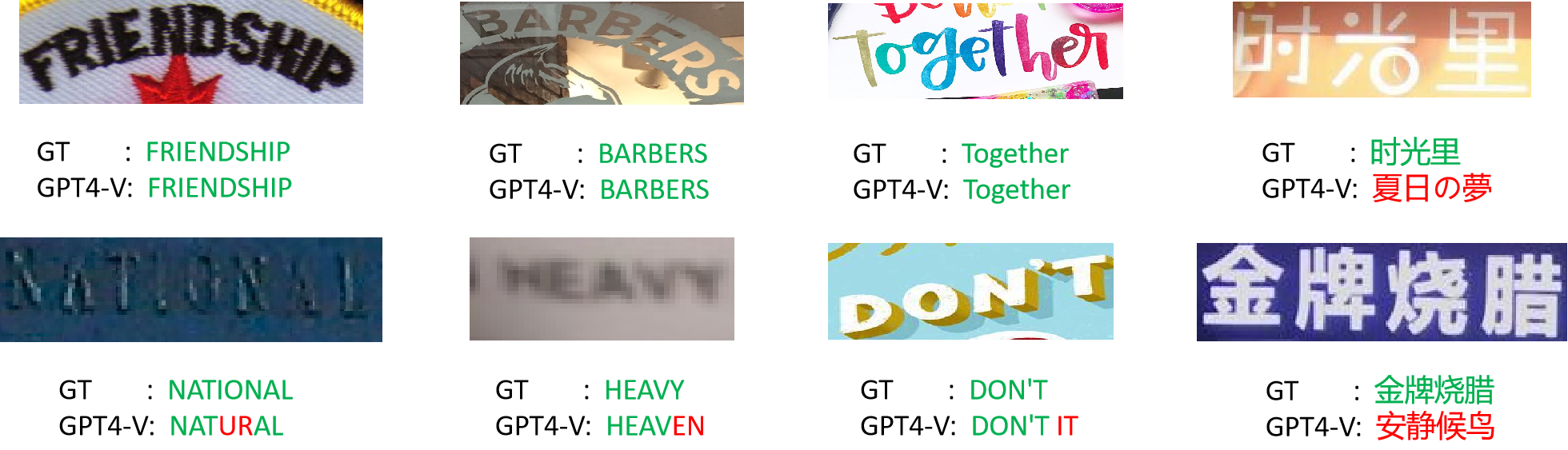}
\caption{Illustration of word-level scene text recognition. In the answers of GPT-4V, we highlight the characters that match the GT in green and characters that do not match in red. GPT-4V can recognize curved, slanted, and artistic English text, while common-style Chinese text can not be recognized.}
\vspace{0em}
\label{fig:vis_STR}
\end{figure*}

\paragraph{Results and analysis} 
The results are shown in Table~\ref{tab:res_STR_word}, Table~\ref{tab:res_STR_e2e} and Table~\ref{tab:res_STR_resize}, respectively. We visualize some examples in Figure~\ref{fig:vis_STR}. Based on the results, we draw the following insights:

\textbf{(1) There is a substantial accuracy disparity between the recognition of English and Chinese text. } As shown in Table~\ref{tab:res_STR_word}, the performance of English text recognition is commendable. Conversely, the accuracy of Chinese text recognition is zero (ReCTS).
We speculate that this may be due to the lack of Chinese scene text images as training data in GPT-4V.

\textbf{(2) GPT-4V exhibits a strong ability to recognize Latin characters, surpassing its performance in other languages.} As shown in Table~\ref{tab:res_STR_e2e}, it can be observed that GPT-4V performs significantly better in English, French, German, and Italian, compared to non-Latin alphabet languages. This suggests noticeable limitations in GPT-4V's multilingual OCR capabilities.

\textbf{(3) GPT-4V supports input images with different resolutions.}
As shown in Table~\ref{tab:res_STR_resize}, there is a positive correlation between the input image resolution and the recognition performance. This suggests that, unlike previous LMMs that resize images to a fixed size, GPT-4V supports input images with variable resolutions.
Meanwhile, we hypothesize that the image encoder of GPT-4V employs a fixed patch size, therefore increasing the resolution of the input image leads to a longer sequence, which help the model to capture more information.


\begin{table*}[h]
\centering
\caption{Results of word-level scene text recognition. The SOTA of CUTE80 and WordArt are achieved by~\cite{jiang2023revisiting} and~\cite{yu2023looking}, respectively.~\cite{yang2022reading} reported the SOTA on SCUT-CTW1500 and Total-Text. The SOTA 
of ReCTS can be found at \protect\footnotemark[3].}
\begin{tabular}{lccccc}
\toprule
Method & CUTE80 & SCUT-CTW1500 & Total-Text & WordArt & ReCTS \\
\midrule
GPT-4V & 88.0\% & 62.0\% & 66.0\% & 62.0\% & 0 \\
Supervised-SOTA & 98.6\% & 87.0\% & 90.1\% & 68.2\% & 97.4\% \\
\bottomrule
\end{tabular}
\label{tab:res_STR_word}
\end{table*}
\footnotetext[3]{\url{https://rrc.cvc.uab.es/?ch=12&com=evaluation&task=2}}



\begin{table*}[h]
\centering
\caption{Results of MLT19. The SOTA 
of end-to-end text spotting in MLT19 can be found at \protect\footnotemark[4].}
\begin{tabular}{llcccc}
\toprule
Method & Language & Precision ↑ & Recall ↑ & F1 ↑ \\
\midrule
\multirow{10}{*}{GPT-4V} & Arabic & 16.44\% & 16.67\% & 16.55\% & \\
 & English & 86.57\% & 78.77\% & 82.49\% & \\
 & French & 83.0\% & 83.84\% & 83.42\% & \\
 & Chinese & 1.2\% & 1.56\% & 1.36\% & \\
 & German & 73.65\% & 86.29\% & 79.47\% & \\
 & Korean & 10.83\% & 12.39\% & 11.56\% & \\
 & Japanese & 11.9\% & 11.9\% & 11.9\% & \\
 & Italian & 62.7\% & 67.52\% & 65.02\% & \\
 & Bangla & 2.53\% & 2.63\% & 2.58\% & \\
 & Hindi & 7.29\% & 8.33\% & 7.78\% & \\
\cmidrule(lr){2-5}
& All language & 43.04\% & 45.42\% & 44.2\% & \\
 \midrule
Supervised-SOTA & All language & 74.16\% & 52.91\% & 61.76\% & \\
\bottomrule
\end{tabular}
\label{tab:res_STR_e2e}
\end{table*}
\footnotetext[4]{\url{https://rrc.cvc.uab.es/?ch=15&com=evaluation&task=4}}


\begin{table*}[h]
\centering
\caption{Impact of image resolution for recognition performance on MLT19 English subset. }
\begin{tabular}{lcccc}
\toprule
Image size & Precision ↑ & Recall ↑ & F1 ↑ \\
\midrule
128  & 45.52\%  &  57.28\%  &  50.73\% \\
256  & 73.88\%  &  86.21\%  &  79.57\% \\
512  & 85.82\%  &  83.21\%  &  84.49\% \\
1024  & 90.30\%  &  84.72\%  &  87.42\% \\
2048  & 92.54\%  &  86.01\%  &  89.16\% \\
\bottomrule
\end{tabular}
\label{tab:res_STR_resize}
\end{table*}

\subsection{Handwritten text recognition}
\paragraph{Dataset} 
To evaluate GPT-4V's capability in handwritten text recognition, we employ two commonly used handwritten datasets: IAM~\cite{marti2002iam} (in English) and CASIA-HWDB~\cite{yin2013icdar} (in Chinese). We randomly sample 50 pages and 50 text lines from each of the test sets of IAM and CASIA-HWDB for evaluation. 

\begin{itemize}
    \item \textbf{IAM} comprises 1,539 pages and 13,353 lines of handwritten English text.
    \item \textbf{CASIA-HWDB} is an offline handwritten Chinese dataset, which contains about 5,090 pages and 1.35 million character samples of 7,356 classes (7,185 Chinese characters and 171 symbols).
\end{itemize}


\paragraph{Prompt} For IAM, we use the prompt: 
\textit{"Recognize the text in the image."} as input.
And for CASIA-HWDB, we use the Chinese prompt 
\begin{CJK*}{UTF8}{gbsn}
"请直接告诉我，图片中的文字都是什么？"
\end{CJK*}
, which means 
\textit{"Please tell me directly, what are all the text in the image?"}
\paragraph{Metric} 


Two metrics are used for evaluation in the handwritten English text: Word Error Rate (WER) and Character Error Rate (CER)~\cite{sueiras2018offline}. 
To evaluate the performance in handwritten Chinese text, we use AR and CR metrics~\cite{peng2022recognition}.


\paragraph{Results and analysis} As shown in Table~\ref{tab:res_HTR_IAM} and ~\ref{tab:res_HTR_CASIA}. 

\textbf{(1) There's also a significant performance gap between English and Chinese handwritten text.} This phenomenon is consistent with the findings in Section~\ref{Sec:STR}, which collectively suggests that GPT-4V performs well in English text recognition while facing notable challenges in Chinese.

\textbf{(2) GPT-4V exhibits significant hallucinations in Chinese text recognition.} As shown in Figure~\ref{fig:vis_HMER} (c) and (d), the responses generated by GPT-4V demonstrate a high degree of fluency in both grammar and semantics. However, they substantially deviate from the textual content of the ground truth (GT), appearing to produce nonsensical information in a seemingly earnest manner.


\begin{table*}[h]
\centering
\caption{Results of IAM. The SOTA of page-level IAM in WER and CER metric are achieved by \cite{kumari2023comprehensive} and \cite{li2023trocr}, respectively. And the line-level SOTA is achieved by \cite{shashank2022improvising}. }
\begin{tabular}{lcccc}
\toprule
\multirow{2}{*}{Method} & \multicolumn{2}{c}{Page-level} & \multicolumn{2}{c}{Line-level} \\
\cmidrule(lr){2-3}  \cmidrule(lr){4-5} 
                        & WER ↓          & CER ↓         & WER ↓          & CER ↓       \\
\midrule
GPT-4V & 9.84\% & 3.32\% & 33.42\% & 13.75\% \\
Supervised-SOTA & 8.29\% & 2.89\% & 21.47\% & 6.52\%  \\
\bottomrule
\end{tabular}
\label{tab:res_HTR_IAM}
\end{table*}

\begin{table*}[h]
\centering
\caption{Results of CASIA-HWDB. The SOTA of page-level CASIA-HWDB in AR and CR metric are achieved by \cite{wang2020writer} and \cite{xie2020high}, respectively. And the line-level SOTA is achieved by \cite{peng2022recognition}.}
\begin{tabular}{lcccc}
\toprule
\multirow{2}{*}{Method} & \multicolumn{2}{c}{Page-level} & \multicolumn{2}{c}{Line-level} \\
\cmidrule(lr){2-3}  \cmidrule(lr){4-5} 
                        & AR ↑          & CR ↑         & AR ↑          & CR ↑         \\
\midrule
GPT-4V & 0.97\% & 36.54\% & -3.45\% & 11.85\% \\
Supervised-SOTA & 96.83\% & 96.99\% & 97.70\% & 97.91\% \\
\bottomrule
\end{tabular}
\label{tab:res_HTR_CASIA}
\end{table*}


\begin{figure*}[h!]
\centering
\includegraphics[width=1\textwidth]{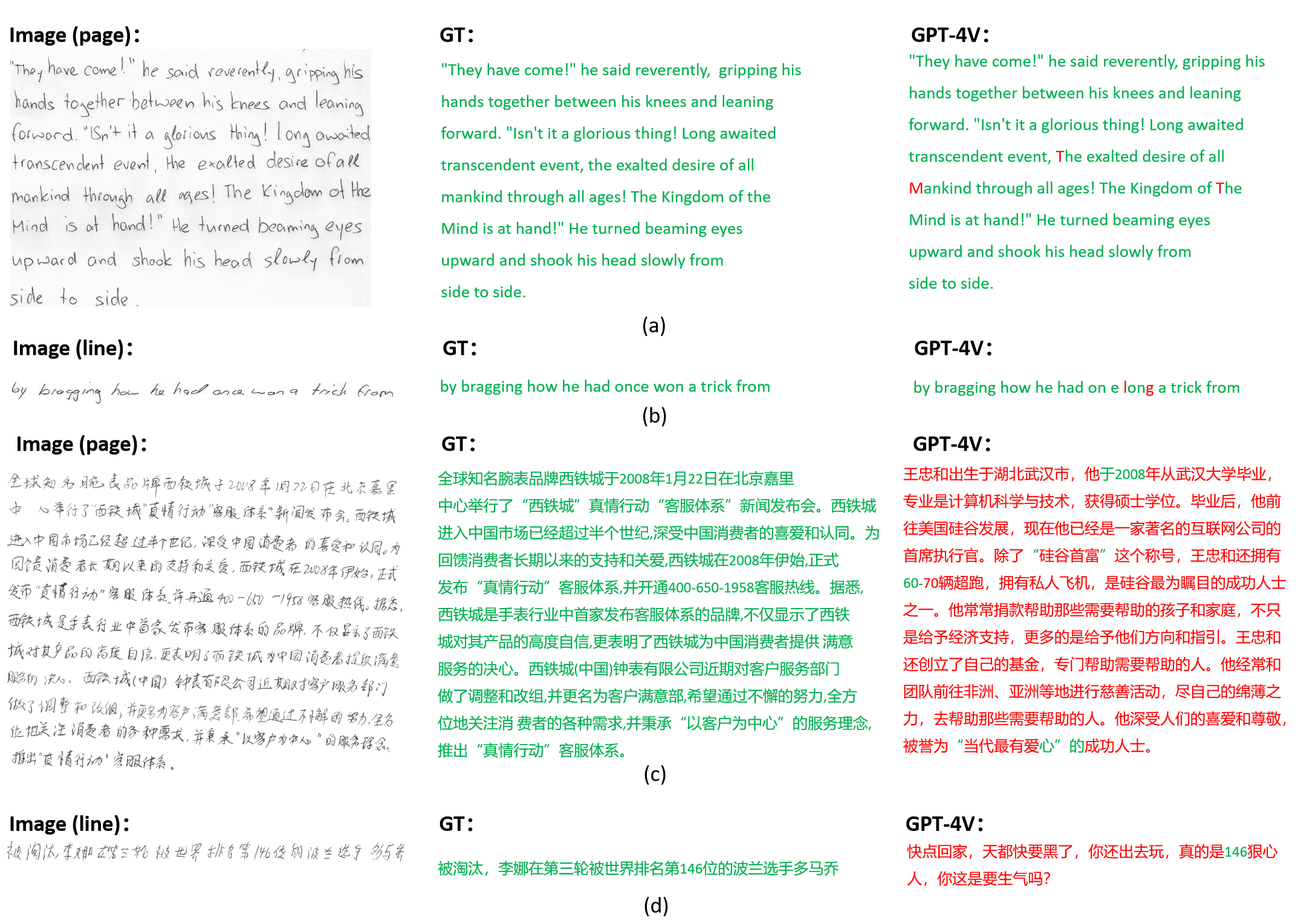}
\caption{Illustration of handwritten text recognition. (a), (b), (c), and (d) are samples of page-level IAM, line-level IAM, page-level CASIA-HWDB, and line-level CASIA-HWDB, respectively. In the responses of GPT-4V, we highlight characters that match the GT in green and characters that do not match in red. For English text, GPT-4V demonstrates excellent performance. In contrast, for Chinese text, GPT-4V has generated a passage of text that is semantically coherent, but it is not associated with the ground truth text (GT). }
\vspace{0em}
\label{fig:vis_CASIA_page}
\end{figure*}

\subsection{Handwritten mathematical expression recognition}
\paragraph{Dataset} For this task, we employ two representative dataset includes CROHME2014~\cite{mouchere2014icfhr} and HME100K~\cite{yuan2022syntax}. We randomly select 50 images from the test sets of each of these two datasets for evaluation.
\begin{itemize}
    \item \textbf{CROHME2014} is a classical online dataset for handwritten mathematical expression recognition, which comprises 9,820 samples of mathematical expressions.
    \item \textbf{HME100K} is a large-scale handwritten mathematical expression recognition dataset, which contains 100k images from ten thousand writers and is mainly captured by cameras. 
\end{itemize}


\paragraph{Prompt} In this task, we use 
\textit{"This is an image of a handwritten mathematical expression. Please recognize the expression above as LaTeX."} as prompt.

\paragraph{Metric} The metrics we employed include the correct rates at the expression level, and with at most one to three errors~\cite{mouchere2014icfhr}.

\paragraph{Results and analysis} The results are shown in Table~\ref{tab:res_HMER}. 
Based on the analysis of the failed case, we draw the following findings.

\begin{figure*}[h!]
\centering
\includegraphics[width=0.95\textwidth]{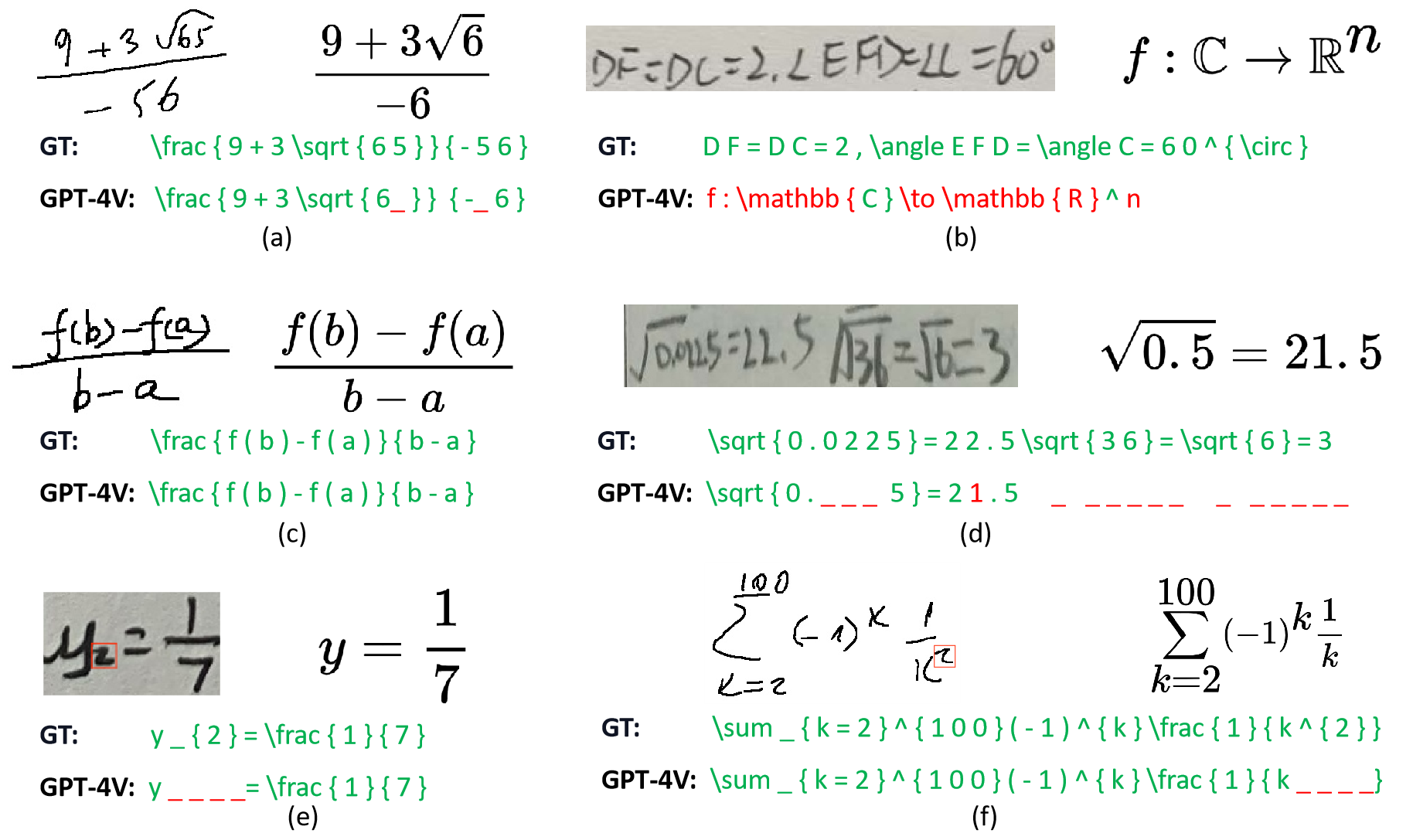}
\caption{Illustration of handwritten mathematical expression recognition. In each example, the left side displays the input image, while the right side shows the image rendered from the LaTeX sequence output by GPT-4V. In the answer of GPT-4V, we highlight elements that match the GT in green and elements that do not match in red. The symbol \_ in red represents the missing elements in the output.}
\vspace{0em}
\label{fig:vis_HMER}
\end{figure*}

\textbf{(1) GPT-4V appears to be limited when dealing with camera-captured and poor handwriting scenarios.} 
As shown in Table~\ref{tab:res_HMER}, the performance on HEM100K (which features camera-captured images and poor handwriting) significantly drops compared to CROHME2014. As shown in Figure~\ref{fig:vis_HMER}, (a) and (c) are examples from CROHME2014, (b) and (d) are from HEM100K, GPT-4V performs well on the former, but poorly on the latter.

\textbf{(2) GPT-4V exhibits certain challenges in fine-grained character recognition.} Among the failed cases, we observed instances where GPT-4V occasionally missed small-scale characters. 
Two examples are shown in Figure~\ref{fig:vis_HMER} (e) and (f). For these two examples, GPT-4V has omitted a superscript and a subscript, respectively.
This finding aligns with the evaluation results of Liu et al.~\cite{liu2023hidden} on other multimodal models, suggesting that GPT-4V may also suffer from certain fine-grained perceptual issues.

\begin{table*}[h]
\centering
\caption{Results of handwritten mathematical expression recognition. The SOTA of CROHME2014 and HME100K are both achieved by~\cite{li2022counting}.}
\begin{tabular}{lcccccccc}
\toprule
\multirow{2}{*}{Method} & \multicolumn{4}{c}{CROHME2014} & \multicolumn{4}{c}{HME100K}  \\
\cmidrule(lr){2-5}  \cmidrule(lr){6-9} 
                        & Exp rate ↑ & <=1 ↑ & <=2 ↑ & <=3 ↑ & Exp rate ↑ & <=1 ↑ & <=2 ↑ & <=3 ↑         \\
\midrule
GPT-4V             & 34.0\% & 44.0\% & 50.0\% & 54.0\% & 16.0\% & 18.0\% & 22.0\% & 28.0\% \\
Supervised-SOTA~\cite{yuan2022syntax} & 65.89\% & 77.97\% & 84.16\% & - & 68.09\% & 83.22\% & 89.91\% & - \\
\bottomrule
\end{tabular}
\label{tab:res_HMER}
\end{table*}

\subsection{Table structure recognition}
\paragraph{Dataset} The datasets we used for this task includes SciTSR~\cite{chi2019complicated} and WTW~\cite{long2021parsing}. 
We randomly select 50 tables from each of the test sets of SciTSR and WTW for evaluation.
Following~\cite{lin2022tsrformer}, we crop table regions from original images for evaluation.
\begin{itemize}
    \item \textbf{SciTSR} is a dedicated dataset created to address the task of table structure recognition in scientific papers. The dataset consists of 12,000 training samples and 3,000 test samples.
    \item \textbf{WTW}'s images are collected in the wild. The dataset is split into training/testing sets with 10,970 and 3,611 samples respectively. 
\end{itemize}

\paragraph{Prompt}For both SciTSR and WTW, we use the prompt 
\textit{"Please read the table in this image 
and return a html-style reconstructed table in text, do not omit anything."} as input.

\paragraph{Metric}To evaluate the performance of GPT-4V in table structure recognition, we use TEDS-S metrics~\cite{zhong2020image}, which 
is a variation of Tree-Edit-Distance-Based Similarity (TEDS)~\cite{zhong2020image} that disregards the 
textual content of the cells and only evaluates the accuracy of the table structure prediction.

\begin{figure*}[h!]
\centering
\includegraphics[width=0.95\textwidth]{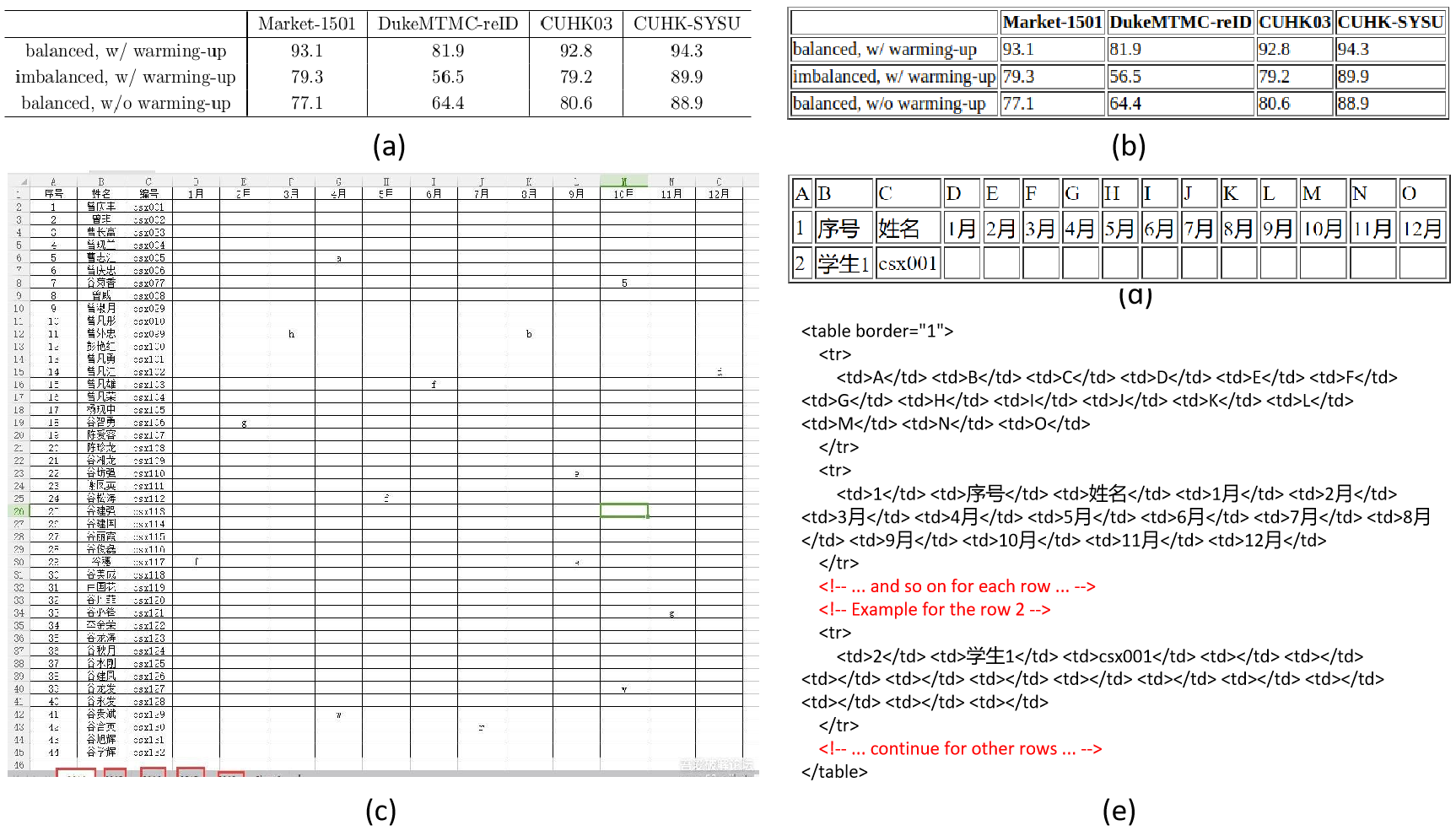}
\caption{Illustration of table structure recognition. (a) and (c) are two input images, (b) and (d) are the corresponding visualized images of GPT-4V's html-style output sequence. (e) is the output sequence of (c), where the elements that GPT-4V indicate the omitted content are highlighted in red.}
\vspace{0em}
\label{fig:vis_TSR}
\end{figure*}

\paragraph{Results and analysis} 
The results are shown in Table~\ref{tab:res_TBR}. We gain two important findings based on the results:

\textbf{(1) GPT-4V struggles with complex tables.} GPT-4V demonstrates outstanding performance when handling tables with structured layouts and consistent text distributions, such as Figure~\ref{fig:vis_TSR} (a). However, when dealing with other types of tables, including those with numerous empty cells, uneven text distribution, skewing, rotation, or densely packed arrangements, its performance noticeably declines.

\textbf{(2) Content omission issues are observed in GPT-4V when processing lengthy tables.} Despite emphasizing the requirement of "do not omit anything" in the prompt, we still observed some instances of content omission in the responses, particularly in the case of a large table.
A typical example is shown in Figure~\ref{fig:vis_TSR} (e), the table image Figure~\ref{fig:vis_TSR} (c) contains many rows, but  GPT-4V only reconstructs three of them.


\begin{table*}[h]
\centering
\caption{The TEDS-S of SciTSR and WTW. The SOTA of SciTSR and WTW are both achieved by~\cite{chen2022complex}.}
\begin{tabular}{lcc}
\toprule
Method & SciTSR & WTW  \\
\midrule
GPT-4V             & 87.47\% & 25.60\%  \\
Supervised-SOTA    & 99.19\% & 91.91\%  \\
\bottomrule
\end{tabular}
\label{tab:res_TBR}
\end{table*}

\subsection{Information Extraction from Visually-rich Document}
\paragraph{Dataset}
We evaluate GPT-4V on FUNSD~\cite{jaume2019funsd} and XFUND~\cite{xu2022xfund} Chinese subset (XFUND-zh). 
\begin{itemize}
    \item \textbf{FUNSD} dataset is a commonly used form understanding benchmark, which contains 199 scanned form-like documents with noisy images.
    \item \textbf{XFUND} dataset is a multilingual extension of FUNSD that covers seven languages (Chinese, Japanese, French, Italian, German, Spanish, and Portuguese).
\end{itemize}

We evaluate GPT-4V on the Semantic Entity Recognition (SER) and the end-to-end Pair Extraction tasks. The SER task requires the model to identify the category of each text segments, which are predefined as header, question, answer, and other in FUNSD and XFUND. The end-to-end pair extraction task asks the model to extract all the key-value pairs in the given document image. We use the full test set (both FUNSD and XFUND-zh contain 50 samples) for performance evaluation.

\paragraph{Prompt}
For FUNSD, we use the following prompt for SER:

\textit{Please read the text in this image and return the information in the following JSON format (note xxx is placeholder, if the information is not available in the image, put "N/A" instead).}
\textit{"header": [xxx, ...], "key": [xxx, ...], "value": [xxx, ...]}

It's important to highlight that, we redefined the official entity type of "\textit{question}" and "\textit{answer}" as "\textit{key}" and "\textit{value}" to maintain consistency with the Pair Extraction task.

For end-to-end Pair Extraction, we use the following prompt:

\textit{You are a document understanding AI, who reads the contents in the given document image and tells the information that the user needs. Respond with the original content in the document image, do not reformat. No extra explanation is needed.
Extract all the key-value pairs from the document image.}

\paragraph{Metric}
For the SER task, we employ the entity-level F1-score \cite{10.1145/3394486.3403172} for performance evaluation. Additionally, Normalized Edit Distance (NED) is also calculated as is done in other end-to-end VIE methods \cite{yu2023structextv}. However, due to limitations in GPT-4V's ability to generate precise bounding boxes for entities, we aligned predictions with ground-truth using the principle of minimum edit distance.

\begin{figure*}[ht]
    \centering
    \includegraphics[width=0.75\textwidth]{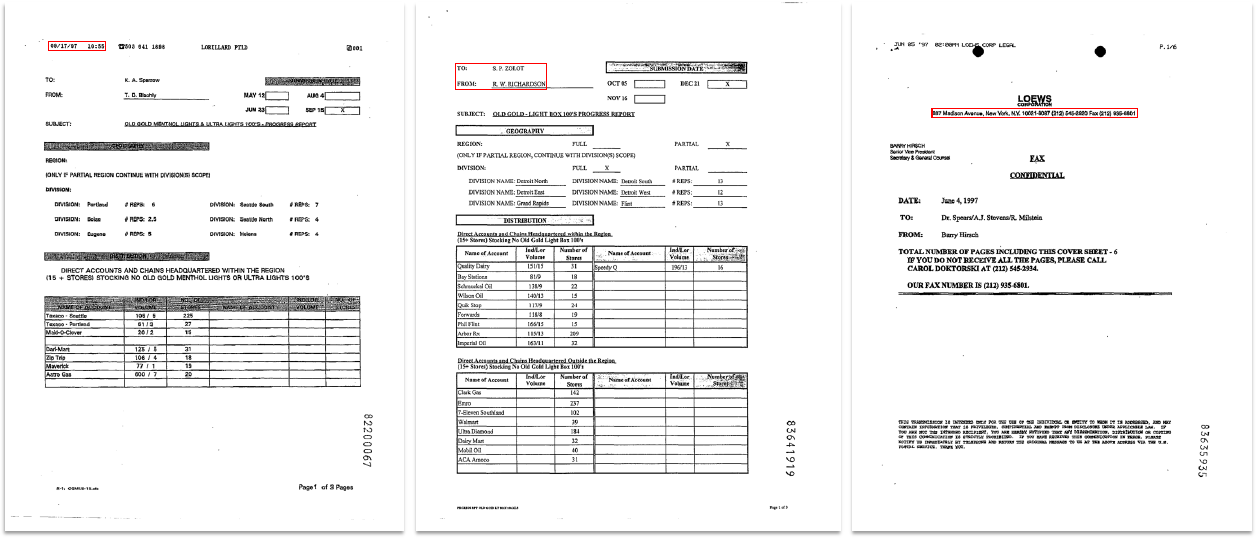}
    \caption{Illustration of error cases of the SER task. The text content enclosed within the red box is incorrectly identified as \textit{header} entities.}
    \label{fig:ser_visualization}
\end{figure*}

\begin{figure*}[!h]
    \centering
    \includegraphics[width=0.95\textwidth]{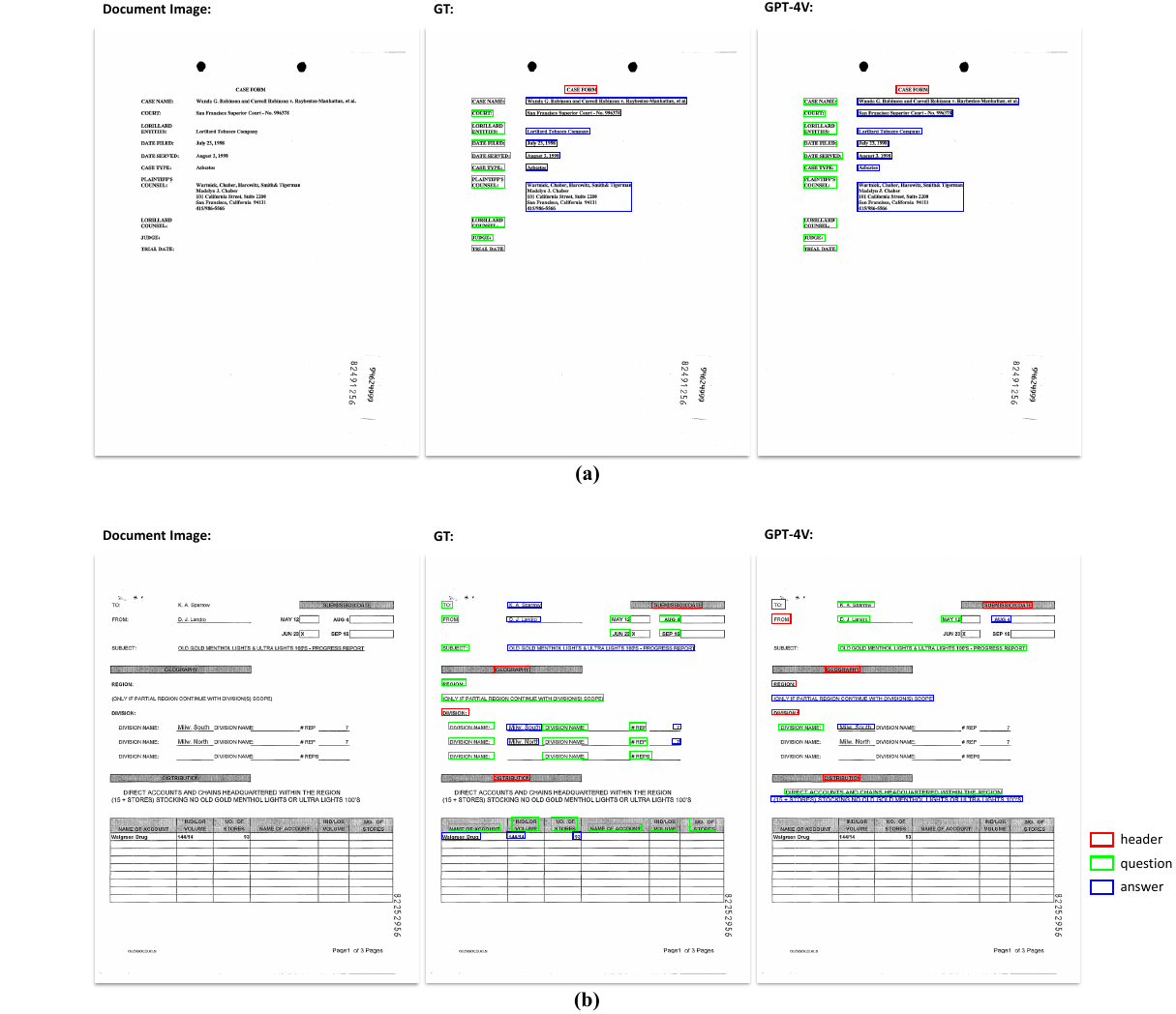}
    \caption{Illustration of Entity Prediction on Full Document Images in the FUNSD Dataset. Due to GPT-4V's limited capability in recognizing Chinese characters, we have excluded examples from the XFUND-zh dataset in this context. Zoom in for the best review.}
    \label{fig:vis_ser}
\end{figure*}

\paragraph{Results and analysis} The SER and Pair Extraction results are shown in \ref{tab:res_ser} and \ref{tab:res_linking}, respectively. We found that: 

\textbf{(1) GPT-4V might have constraints in comprehending the spatial arrangement of documents.} As shown in Figure \ref{fig:ser_visualization}, some text content located near the top of the page, which lacks both visual and semantic alignment with the \textit{header} category, is erroneously identified as a \textit{header}. Additional visualizations are presented in \ref{fig:vis_ser}. It is evident that GPT-4V excels in analyzing documents with straightforward layouts but struggles to comprehend those featuring intricate layouts. 

\begin{figure*}[h]
    \centering
    \includegraphics[width=0.5\textwidth]{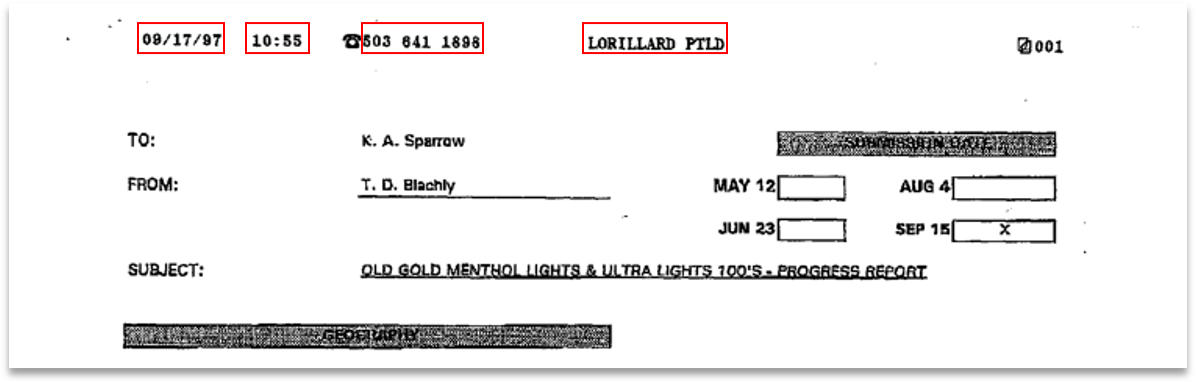}
    \caption{Illustration of error cases of the Pair Extraction task. The text content enclosed within the red box is incorrectly identified as entity pairs.}
    \label{fig:linking_visualization}
\end{figure*}

\textbf{(2) GPT4V tends to generate new keys for non-kv-pair contents.} For instance, as shown in Figure \ref{fig:linking_visualization}, contents "09 / 17 / 97  10:55" at the header part are recognized as "Date: 09/18/97", "Time: 10:55", "Fax Number: 503 841 1898", "Company: LORILLARD PTLD", "Page Number: 001". 

\begin{table*}[h]
\centering
\caption{SER Results of FUNSD and XFUND-zh. The SOTA of FUNSD is provided in \cite{yu2023structextv}.}
\resizebox{0.95\textwidth}{!}{
\begin{tabular}{lcccccccc}
\toprule
\multirow{2}{*}{Method} & \multicolumn{4}{c}{FUNSD} & \multicolumn{4}{c}{XFUND-zh}  \\
\cmidrule(lr){2-5}  \cmidrule(lr){6-9} 
                        & Precision ↑ & Recall ↑ & F1 ↑ & 1-NED ↑ & Precision ↑ & Recall ↑ & F1 ↑ & 1-NED ↑        \\
\midrule
GPT-4V             & 41.85\% & 29.36\% & 34.51\% & 0.2697 & 25.87\% & 15.15\% & 19.11\% & 0.1544 \\
Supervised-SOTA \cite{yu2023structextv} & - & - & - & 0.5500 & - & - & - & - \\
\bottomrule
\end{tabular}
}
\label{tab:res_ser}
\end{table*}

\begin{table*}[ht]
\centering
\caption{Pair Extraction Results of FUNSD and XFUND-zh.}
\begin{tabular}{ccccc}
\toprule
Dataset          & Precision ↑   & Recall ↑  & F1 ↑  & 1-NED ↑  \\
\midrule
FUNSD            & 20.69\%     & 10.25\%     & 13.71\%     & 0.1979    \\
XFUND-zh         & 0.07\%     & 0.02\%     & 0.03\%     & 0.0420    \\
\bottomrule
\end{tabular}
\label{tab:res_linking}
\end{table*}

\section{Discussions}
\label{discuss}
\paragraph{Do specialized models still hold research value in the OCR field?} 
There are three main drawbacks of GPT-4V.
(1) Based on the experimental results in Section~\ref{Experiments}, GPT-4V's ability in OCR is limited to Latin contents and struggles to cope with multilingual and complex scenarios.
(2) The inference cost and delay are significantly high, thereby posing usability challenges in some practical scenarios.
(3) The long cycle and complex process of updating make it difficult to promptly address minor issues.
Considering the aforementioned shortcomings and limited OCR capabilities of some other LMMs~\cite{liu2023hidden}, we believe that existing LMMs struggle to simultaneously excel in various OCR tasks. 
Therefore, \textit{\textbf{we contend that specialized models in the field of OCR continue to hold significant value for research. }}

\paragraph{How can we fully leverage the potential of LMMs like GPT-4V in the OCR domain?} 
These are some possible strategies.
\textbf{(1) Semantic understanding enhancement:}
A significant characteristic of LMMs lies in their outstanding semantic capabilities after extensive training on large-scale data. Since semantic understanding is a crucial factor in document comprehension and some related tasks, harnessing the semantic potential of LMMs can greatly enhance the performance in these tasks.
\textbf{(2) Downstream task finetuning:} Another approach that fully leverages the prior knowledge of LMMs is fine-tuning, especially in scenarios with limited data. Fine-tuning allows the model to adapt to specific tasks or domains, thus improving the performance~\cite{instructblip}.
\textbf{(3) Auto/semi-auto data construction:} Using LMMs for automatic/semi-automatic data annotation and generation will substantially reduce the cost of manual labeling, which is an effective strategy for tackling the difficulties of data acquisition~\cite{wei2023instructiongpt}.

\section{Limitations}
There are three main limitations of our work.
First, the test sample of our evaluation is small-scale (mostly 50 samples per dataset) due to the conversation limits (50 conversations per 3 hours) of GPT-4V. This could potentially limit the generalizability of the results.
Second, our assessment primarily focuses on mainstream OCR tasks and does not include other OCR-related tasks. Hence, the findings might not cover the full spectrum of OCR capabilities of GPT-4V.
Third, only the zero-shot capacity of GPT-4V in OCR was evaluated, without exploring few-shot scenarios.
As a result, the potential benefits of further training or fine-tuning the LLM model for specific tasks are not addressed. Few-shot scenarios with technology such as in-context learning~\cite{brown2020language} are worth of exploring in the future.

\section{Conclusion}
In this paper, we present a comprehensive evaluation of the OCR capabilities of GPT-4V through a variety of experiments. For the first time, we offer not only qualitative demonstrations but also quantitative performance analysis of GPT-4V across a wide spectrum of tasks. These tasks encompass scene text recognition, handwritten text recognition, handwritten mathematical expression recognition, table structure recognition, and information extraction from visually rich documents.

Our findings, grounded in meticulous experimental results, provide an in-depth analysis of the strengths and limitations of GPT-4V. 
Although the model shows a strong ability to accurately recognize Latin content and supports input images of variable resolutions, it displays notable struggles with multilingual and complex scenarios. Additionally, the high inference costs and the challenges associated with continuous updating pose significant barriers to the real-world deployment of GPT-4V. 
Therefore, we contend that specialized models in the field of OCR continue to hold significant value for research.
Despite these limitations, GPT-4V and other existing general LMMs could still significantly contribute to the development of the OCR field in several ways. These would include enhancing semantic understanding, fine-tuning for downstream tasks, and facilitating auto/semi-auto data construction.

In summary, this paper presents a first-of-its-kind, in-depth quantitative evaluation of GPT-4V's performance in OCR tasks. We will continuously update the evaluation results in the future, and we hope the findings in this paper will provide valuable insights and strategies for researchers 
and practitioners working on OCR tasks using large multi-modal models. 

\bibliographystyle{unsrt}  
\bibliography{references}

\begin{thebibliography}{10}

\bibitem{chatgpt}
OpenAI.
\newblock Chat{GPT}.
\newblock \url{https://openai.com/blog/chatgpt/}, 2023.

\bibitem{zeng2022glm}
Aohan Zeng, Xiao Liu, Zhengxiao Du, Zihan Wang, Hanyu Lai, Ming Ding, Zhuoyi Yang, Yifan Xu, Wendi Zheng, Xiao Xia, et~al.
\newblock {GLM-130B}: An open bilingual pre-trained model.
\newblock {\em arXiv preprint arXiv:2210.02414}, 2022.

\bibitem{alpaca}
Rohan Taori, Ishaan Gulrajani, Tianyi Zhang, Yann Dubois, Xuechen Li, Carlos Guestrin, Percy Liang, and Tatsunori~B. Hashimoto.
\newblock Stanford {Alpaca}: An instruction-following {LLaMA} model.
\newblock \url{https://github.com/tatsu-lab/stanford_alpaca}, 2023.

\bibitem{chiang2023vicuna}
Wei-Lin Chiang, Zhuohan Li, Zi~Lin, Ying Sheng, Zhanghao Wu, Hao Zhang, Lianmin Zheng, Siyuan Zhuang, Yonghao Zhuang, Joseph~E Gonzalez, et~al.
\newblock {Vicuna}: An open-source chatbot impressing {GPT}-4 with 90\%* {ChatGPT} quality.
\newblock {\em See https://vicuna. lmsys. org (accessed 14 April 2023)}, 2023.

\bibitem{touvron2023llama}
Hugo Touvron, Thibaut Lavril, Gautier Izacard, Xavier Martinet, Marie-Anne Lachaux, Timoth{\'e}e Lacroix, Baptiste Rozi{\`e}re, Naman Goyal, Eric Hambro, Faisal Azhar, et~al.
\newblock {LLaMA}: Open and efficient foundation language models.
\newblock {\em arXiv preprint arXiv:2302.13971}, 2023.

\bibitem{wenxin}
Baidu.
\newblock {ERNIE Bot}.
\newblock \url{https://yiyan.baidu.com/}, 2023.

\bibitem{tongyi}
Ali.
\newblock Qwen.
\newblock \url{https://tongyi.aliyun.com/}, 2023.

\bibitem{baichuan2023baichuan2}
Baichuan.
\newblock Baichuan 2: Open large-scale language models.
\newblock {\em arXiv preprint arXiv:2309.10305}, 2023.

\bibitem{li2023blip2}
Junnan Li, Dongxu Li, Silvio Savarese, and Steven Hoi.
\newblock {BLIP-2}: Bootstrapping language-image pre-training with frozen image encoders and large language models, 2023.

\bibitem{anas_awadalla_2023_7733589}
Anas Awadalla, Irena Gao, Joshua Gardner, Jack Hessel, Yusuf Hanafy, Wanrong Zhu, Kalyani Marathe, Yonatan Bitton, Samir Gadre, Jenia Jitsev, Simon Kornblith, Pang~Wei Koh, Gabriel Ilharco, Mitchell Wortsman, and Ludwig Schmidt.
\newblock Openflamingo, March 2023.

\bibitem{liu2023visual}
Haotian Liu, Chunyuan Li, Qingyang Wu, and Yong~Jae Lee.
\newblock Visual instruction tuning.
\newblock {\em arXiv preprint arXiv:2304.08485}, 2023.

\bibitem{zhu2023minigpt4}
Deyao Zhu, Jun Chen, Xiaoqian Shen, Xiang Li, and Mohamed Elhoseiny.
\newblock {MiniGPT-4}: Enhancing vision-language understanding with advanced large language models, 2023.

\bibitem{ye2023mplugowl}
Qinghao Ye, Haiyang Xu, Guohai Xu, Jiabo Ye, Ming Yan, Yiyang Zhou, Junyang Wang, Anwen Hu, Pengcheng Shi, Yaya Shi, Chenliang Li, Yuanhong Xu, Hehong Chen, Junfeng Tian, Qian Qi, Ji~Zhang, and Fei Huang.
\newblock {mPLUG-Owl}: Modularization empowers large language models with multimodality, 2023.

\bibitem{yang2023dawn}
Zhengyuan Yang, Linjie Li, Kevin Lin, Jianfeng Wang, Chung-Ching Lin, Zicheng Liu, and Lijuan Wang.
\newblock The dawn of {LMMs}: Preliminary explorations with {GPT-4V} (ision).
\newblock {\em arXiv preprint arXiv:2309.17421}, 2023.

\bibitem{shi2016end}
Baoguang Shi, Xiang Bai, and Cong Yao.
\newblock An end-to-end trainable neural network for image-based sequence recognition and its application to scene text recognition.
\newblock {\em {IEEE} Trans. Pattern Anal. Mach. Intell.}, 39(11):2298--2304, 2016.

\bibitem{shi2018aster}
Baoguang Shi, Mingkun Yang, Xinggang Wang, Pengyuan Lyu, Cong Yao, and Xiang Bai.
\newblock Aster: An attentional scene text recognizer with flexible rectification.
\newblock {\em {IEEE} Trans. Pattern Anal. Mach. Intell.}, 41(9):2035--2048, 2018.

\bibitem{luo2019moran}
Canjie Luo, Lianwen Jin, and Zenghui Sun.
\newblock Moran: A multi-object rectified attention network for scene text recognition.
\newblock {\em Pattern Recognit.}, 90:109--118, 2019.

\bibitem{wang2020decoupled}
Tianwei Wang, Yuanzhi Zhu, Lianwen Jin, Canjie Luo, Xiaoxue Chen, Yaqiang Wu, Qianying Wang, and Mingxiang Cai.
\newblock Decoupled attention network for text recognition.
\newblock In {\em Proc. AAAI}, volume~34, pages 12216--12224, 2020.

\bibitem{liao2020mask}
Minghui Liao, Guan Pang, Jing Huang, Tal Hassner, and Xiang Bai.
\newblock Mask {TextSpotter} v3: Segmentation proposal network for robust scene text spotting.
\newblock In {\em Proc. ECCV}, pages 706--722. Springer, 2020.

\bibitem{fang2021read}
Shancheng Fang, Hongtao Xie, Yuxin Wang, Zhendong Mao, and Yongdong Zhang.
\newblock Read like humans: Autonomous, bidirectional and iterative language modeling for scene text recognition.
\newblock In {\em Proc. CVPR}, pages 7098--7107, 2021.

\bibitem{liu2021abcnet}
Yuliang Liu, Chunhua Shen, Lianwen Jin, Tong He, Peng Chen, Chongyu Liu, and Hao Chen.
\newblock {ABCNet} v2: Adaptive bezier-curve network for real-time end-to-end text spotting.
\newblock {\em {IEEE} Trans. Pattern Anal. Mach. Intell.}, 44(11):8048--8064, 2021.

\bibitem{zhong2022sgbanet}
Dajian Zhong, Shujing Lyu, Palaiahnakote Shivakumara, Bing Yin, Jiajia Wu, Umapada Pal, and Yue Lu.
\newblock {SGBANet}: semantic gan and balanced attention network for arbitrarily oriented scene text recognition.
\newblock In {\em Proc. ECCV}, pages 464--480. Springer, 2022.

\bibitem{huang2022swintextspotter}
Mingxin Huang, Yuliang Liu, Zhenghao Peng, Chongyu Liu, Dahua Lin, Shenggao Zhu, Nicholas Yuan, Kai Ding, and Lianwen Jin.
\newblock {SwinTextSpotter}: Scene text spotting via better synergy between text detection and text recognition.
\newblock In {\em Proc. CVPR}, pages 4593--4603, 2022.

\bibitem{ye2023deepsolo}
Maoyuan Ye, Jing Zhang, Shanshan Zhao, Juhua Liu, Tongliang Liu, Bo~Du, and Dacheng Tao.
\newblock {DeepSolo}: Let transformer decoder with explicit points solo for text spotting.
\newblock In {\em Proc. CVPR}, pages 19348--19357, 2023.

\bibitem{huang2023estextspotter}
Mingxin Huang, Jiaxin Zhang, Dezhi Peng, Hao Lu, Can Huang, Yuliang Liu, Xiang Bai, and Lianwen Jin.
\newblock {ESTextSpotter}: Towards better scene text spotting with explicit synergy in transformer.
\newblock In {\em Proc. ICCV}, pages 19495--19505, 2023.

\bibitem{lyu2018mask}
Pengyuan Lyu, Minghui Liao, Cong Yao, Wenhao Wu, and Xiang Bai.
\newblock Mask {TextSpotter}: An end-to-end trainable neural network for spotting text with arbitrary shapes.
\newblock In {\em Proc. ECCV}, pages 67--83, 2018.

\bibitem{liao2019mask}
Minghui Liao, Pengyuan Lyu, Minghang He, Cong Yao, Wenhao Wu, and Xiang Bai.
\newblock {Mask TextSpotter}: An end-to-end trainable neural network for spotting text with arbitrary shapes.
\newblock {\em {IEEE} Trans. Pattern Anal. Mach. Intell.}, 43(2):532--548, 2019.

\bibitem{peng2022spts}
Dezhi Peng, Xinyu Wang, Yuliang Liu, Jiaxin Zhang, Mingxin Huang, Songxuan Lai, Jing Li, Shenggao Zhu, Dahua Lin, Chunhua Shen, et~al.
\newblock {SPTS}: single-point text spotting.
\newblock In {\em Proc. ACM MM}, pages 4272--4281, 2022.

\bibitem{liu2023sptsv2}
Yuliang Liu, Jiaxin Zhang, Dezhi Peng, Mingxin Huang, Xinyu Wang, Jingqun Tang, Can Huang, Dahua Lin, Chunhua Shen, Xiang Bai, and Lianwen Jin.
\newblock {SPTS v2}: Single-point scene text spotting.
\newblock {\em {IEEE} Trans. Pattern Anal. Mach. Intell.}, pages 1--15, 2023.

\bibitem{zhang2023arbitrary}
Shi-Xue Zhang, Chun Yang, Xiaobin Zhu, and Xu-Cheng Yin.
\newblock Arbitrary shape text detection via boundary transformer.
\newblock {\em {IEEE} Trans. Multimedia}, 2023.

\bibitem{wang2011handwritten}
Qiu-Feng Wang, Fei Yin, and Cheng-Lin Liu.
\newblock Handwritten {Chinese} text recognition by integrating multiple contexts.
\newblock {\em {IEEE} Trans. Pattern Anal. Mach. Intell.}, 34(8):1469--1481, 2011.

\bibitem{bluche2016joint}
Th{\'e}odore Bluche.
\newblock Joint line segmentation and transcription for end-to-end handwritten paragraph recognition.
\newblock {\em Proc. NIPS}, 29, 2016.

\bibitem{xie2017learning}
Zecheng Xie, Zenghui Sun, Lianwen Jin, Hao Ni, and Terry Lyons.
\newblock Learning spatial-semantic context with fully convolutional recurrent network for online handwritten {Chinese} text recognition.
\newblock {\em {IEEE} Trans. Pattern Anal. Mach. Intell.}, 40(8):1903--1917, 2017.

\bibitem{peng2019fast}
Dezhi Peng, Lianwen Jin, Yaqiang Wu, Zhepeng Wang, and Mingxiang Cai.
\newblock A fast and accurate fully convolutional network for end-to-end handwritten {Chinese} text segmentation and recognition.
\newblock In {\em Proc. ICDAR}, pages 25--30. IEEE, 2019.

\bibitem{yousef2020origaminet}
Mohamed Yousef and Tom~E Bishop.
\newblock {OrigamiNet}: weakly-supervised, segmentation-free, one-step, full page text recognition by learning to unfold.
\newblock In {\em Proc. CVPR}, pages 14710--14719, 2020.

\bibitem{peng2022recognition}
Dezhi Peng, Lianwen Jin, Weihong Ma, Canyu Xie, Hesuo Zhang, Shenggao Zhu, and Jing Li.
\newblock Recognition of handwritten {Chinese} text by segmentation: a segment-annotation-free approach.
\newblock {\em {IEEE} Trans. Multimedia}, 2022.

\bibitem{peng2022pagenet}
Dezhi Peng, Lianwen Jin, Yuliang Liu, Canjie Luo, and Songxuan Lai.
\newblock {PageNet}: Towards end-to-end weakly supervised page-level handwritten {Chinese} text recognition.
\newblock {\em Int. J. Comput. Vis.}, 130(11):2623--2645, 2022.

\bibitem{huang2023segctc}
Jiarong Huang, Dezhi Peng, Hongliang Li, Hao Ni, and Lianwen Jin.
\newblock {SegCTC}: Offline handwritten {Chinese} text recognition via better fusion between explicit and implicit segmentation.
\newblock In {\em Proc. ICDAR}, pages 332--349. Springer, 2023.

\bibitem{coquenet2023dan}
Denis Coquenet, Cl{\'e}ment Chatelain, and Thierry Paquet.
\newblock {DAN}: a segmentation-free document attention network for handwritten document recognition.
\newblock {\em {IEEE} Trans. Pattern Anal. Mach. Intell.}, 2023.

\bibitem{zhang2017watch}
Jianshu Zhang, Jun Du, Shiliang Zhang, Dan Liu, Yulong Hu, Jinshui Hu, Si~Wei, and Lirong Dai.
\newblock Watch, attend and parse: An end-to-end neural network based approach to handwritten mathematical expression recognition.
\newblock {\em Pattern Recognit.}, 71:196--206, 2017.

\bibitem{zhang2018multi}
Jianshu Zhang, Jun Du, and Lirong Dai.
\newblock Multi-scale attention with dense encoder for handwritten mathematical expression recognition.
\newblock In {\em Proc. ICPR}, pages 2245--2250. IEEE, 2018.

\bibitem{yuan2022syntax}
Ye~Yuan, Xiao Liu, Wondimu Dikubab, Hui Liu, Zhilong Ji, Zhongqin Wu, and Xiang Bai.
\newblock Syntax-aware network for handwritten mathematical expression recognition.
\newblock In {\em Proc. CVPR}, pages 4553--4562, 2022.

\bibitem{zhong2022tree}
Shuhan Zhong, Sizhe Song, Guanyao Li, and S-H~Gary Chan.
\newblock A tree-based structure-aware transformer decoder for image-to-markup generation.
\newblock In {\em Proc. ACM MM}, pages 5751--5760, 2022.

\bibitem{bian2022handwritten}
Xiaohang Bian, Bo~Qin, Xiaozhe Xin, Jianwu Li, Xuefeng Su, and Yanfeng Wang.
\newblock Handwritten mathematical expression recognition via attention aggregation based bi-directional mutual learning.
\newblock In {\em Proc. AAAI}, volume~36, pages 113--121, 2022.

\bibitem{zhao2022comer}
Wenqi Zhao and Liangcai Gao.
\newblock Comer: Modeling coverage for transformer-based handwritten mathematical expression recognition.
\newblock In {\em Proc. ECCV}, pages 392--408. Springer, 2022.

\bibitem{li2022counting}
Bohan Li, Ye~Yuan, Dingkang Liang, Xiao Liu, Zhilong Ji, Jinfeng Bai, Wenyu Liu, and Xiang Bai.
\newblock When counting meets {HMER}: Counting-aware network for handwritten mathematical expression recognition.
\newblock In {\em Proc. ECCV}, pages 197--214. Springer, 2022.

\bibitem{li2023improving}
Zhe Li, Xinyu Wang, Yuliang Liu, Lianwen Jin, Yichao Huang, and Kai Ding.
\newblock Improving handwritten mathematical expression recognition via similar symbol distinguishing.
\newblock {\em {IEEE} Trans. Multimedia}, 2023.

\bibitem{zhong2020image}
Xu~Zhong, Elaheh ShafieiBavani, and Antonio Jimeno~Yepes.
\newblock Image-based table recognition: data, model, and evaluation.
\newblock In {\em Proc. ECCV}, pages 564--580. Springer, 2020.

\bibitem{ye2021pingan}
Jiaquan Ye, Xianbiao Qi, Yelin He, Yihao Chen, Dengyi Gu, Peng Gao, and Rong Xiao.
\newblock Pingan-vcgroup's solution for {ICDAR} 2021 competition on scientific literature parsing task b: table recognition to html.
\newblock {\em arXiv preprint arXiv:2105.01848}, 2021.

\bibitem{liu2021show}
Hao Liu, Xin Li, Bing Liu, Deqiang Jiang, Yinsong Liu, Bo~Ren, and Rongrong Ji.
\newblock Show, read and reason: Table structure recognition with flexible context aggregator.
\newblock In {\em Proc. ACM MM}, pages 1084--1092, 2021.

\bibitem{liu2022neural}
Hao Liu, Xin Li, Bing Liu, Deqiang Jiang, Yinsong Liu, and Bo~Ren.
\newblock Neural collaborative graph machines for table structure recognition.
\newblock In {\em Proc. CVPR}, pages 4533--4542, 2022.

\bibitem{chen2022complex}
Bangdong Chen, Dezhi Peng, Jiaxin Zhang, Yujin Ren, and Lianwen Jin.
\newblock Complex table structure recognition in the wild using transformer and identity matrix-based augmentation.
\newblock In {\em Proc. ICFHR}, pages 545--561. Springer, 2022.

\bibitem{lin2022tsrformer}
Weihong Lin, Zheng Sun, Chixiang Ma, Mingze Li, Jiawei Wang, Lei Sun, and Qiang Huo.
\newblock {TSRFormer}: Table structure recognition with transformers.
\newblock In {\em Proc. ACM MM}, pages 6473--6482, 2022.

\bibitem{huang2023improving}
Yongshuai Huang, Ning Lu, Dapeng Chen, Yibo Li, Zecheng Xie, Shenggao Zhu, Liangcai Gao, and Wei Peng.
\newblock Improving table structure recognition with visual-alignment sequential coordinate modeling.
\newblock In {\em Proc. CVPR}, pages 11134--11143, 2023.

\bibitem{shen2023divide}
Huawen Shen, Xiang Gao, Jin Wei, Liang Qiao, Yu~Zhou, Qiang Li, and Zhanzhan Cheng.
\newblock Divide rows and conquer cells: Towards structure recognition for large tables.
\newblock IJCAI, 2023.

\bibitem{9412927}
Wenwen Yu, Ning Lu, Xianbiao Qi, Ping Gong, and Rong Xiao.
\newblock {PICK}: Processing key information extraction from documents using improved graph learning-convolutional networks.
\newblock In {\em Proc. ICPR}, pages 4363--4370, 2021.

\bibitem{wang-etal-2022-lilt}
Jiapeng Wang, Lianwen Jin, and Kai Ding.
\newblock {L}i{LT}: A simple yet effective language-independent layout transformer for structured document understanding.
\newblock In {\em Proc. ACL}, pages 7747--7757, 2022.

\bibitem{Wang_Liu_Jin_Tang_Zhang_Zhang_Wang_Wu_Cai_2021}
Jiapeng Wang, Chongyu Liu, Lianwen Jin, Guozhi Tang, Jiaxin Zhang, Shuaitao Zhang, Qianying Wang, Yaqiang Wu, and Mingxiang Cai.
\newblock Towards robust visual information extraction in real world: New dataset and novel solution.
\newblock 35:2738--2745, 2021.

\bibitem{DBLP:conf/ijcai/WangWTJMDH21}
Jiapeng Wang, Tianwei Wang, Guozhi Tang, Lianwen Jin, Weihong Ma, Kai Ding, and Yichao Huang.
\newblock Tag, copy or predict: {A} unified weakly-supervised learning framework for visual information extraction using sequences.
\newblock In {\em Proc. IJCAI}, pages 1082--1090, 2021.

\bibitem{10.1145/3394486.3403172}
Yiheng Xu, Minghao Li, Lei Cui, Shaohan Huang, Furu Wei, and Ming Zhou.
\newblock {LayoutLM}: Pre-training of text and layout for document image understanding.
\newblock In {\em Proc. SIGKDD}, page 1192–1200, 2020.

\bibitem{10.1145/3503161.3548112}
Yupan Huang, Tengchao Lv, Lei Cui, Yutong Lu, and Furu Wei.
\newblock {LayoutLMv3}: Pre-training for document ai with unified text and image masking.
\newblock In {\em Proc. ACM MM}, page 4083–4091, 2022.

\bibitem{DBLP:conf/aaai/HongKJHNP22}
Teakgyu Hong, Donghyun Kim, Mingi Ji, Wonseok Hwang, Daehyun Nam, and Sungrae Park.
\newblock {BROS:} {A} pre-trained language model focusing on text and layout for better key information extraction from documents.
\newblock In {\em Proc. AAAI}, pages 10767--10775, 2022.

\bibitem{Li_2021_CVPR}
Peizhao Li, Jiuxiang Gu, Jason Kuen, Vlad~I. Morariu, Handong Zhao, Rajiv Jain, Varun Manjunatha, and Hongfu Liu.
\newblock {SelfDoc}: Self-supervised document representation learning.
\newblock In {\em Proc. CVPR}, pages 5652--5660, June 2021.

\bibitem{Appalaraju_2021_ICCV}
Srikar Appalaraju, Bhavan Jasani, Bhargava~Urala Kota, Yusheng Xie, and R.~Manmatha.
\newblock {DocFormer}: End-to-end transformer for document understanding.
\newblock In {\em Proc. ICCV}, pages 993--1003, October 2021.

\bibitem{yu2023structextv}
Yuechen Yu, Yulin Li, Chengquan Zhang, Xiaoqiang Zhang, Zengyuan Guo, Xiameng Qin, Kun Yao, Junyu Han, Errui Ding, and Jingdong Wang.
\newblock {StrucTexTv2}: Masked visual-textual prediction for document image pre-training.
\newblock In {\em Proc. ICLR}, 2023.

\bibitem{Risnumawan2014ARA}
Anhar Risnumawan, Palaiahnakote Shivakumara, Chee~Seng Chan, and Chew~Lim Tan.
\newblock A robust arbitrary text detection system for natural scene images.
\newblock {\em Expert Syst. Appl.}, 41:8027--8048, 2014.

\bibitem{Liu2019CurvedST}
Yuliang Liu, Lianwen Jin, Shuaitao Zhang, Canjie Luo, and Sheng Zhang.
\newblock Curved scene text detection via transverse and longitudinal sequence connection.
\newblock {\em Pattern Recognit.}, 90:337--345, 2019.

\bibitem{Chng2017TotalTextAC}
Chee-Kheng Chng and Chee~Seng Chan.
\newblock {Total-Text}: A comprehensive dataset for scene text detection and recognition.
\newblock In {\em Proc. ICDAR}, volume~01, pages 935--942, 2017.

\bibitem{xie2022toward}
Xudong Xie, Ling Fu, Zhifei Zhang, Zhaowen Wang, and Xiang Bai.
\newblock Toward understanding {WordArt}: Corner-guided transformer for scene text recognition.
\newblock In {\em Proc. ECCV}, pages 303--321. Springer, 2022.

\bibitem{zhang2019icdar}
Rui Zhang, Yongsheng Zhou, Qianyi Jiang, Qi~Song, Nan Li, Kai Zhou, Lei Wang, Dong Wang, Minghui Liao, Mingkun Yang, et~al.
\newblock {ICDAR} 2019 robust reading challenge on reading {Chinese} text on signboard.
\newblock In {\em Proc. ICDAR}, pages 1577--1581. IEEE, 2019.

\bibitem{nayef2019icdar2019}
Nibal Nayef, Yash Patel, Michal Busta, Pinaki~Nath Chowdhury, Dimosthenis Karatzas, Wafa Khlif, Jiri Matas, Umapada Pal, Jean-Christophe Burie, Cheng-lin Liu, et~al.
\newblock {ICDAR2019} robust reading challenge on multi-lingual scene text detection and recognition{—RRC-MLT}-2019.
\newblock In {\em Proc. ICDAR}, pages 1582--1587. IEEE, 2019.

\bibitem{marti2002iam}
U-V Marti and Horst Bunke.
\newblock The {IAM}-database: an english sentence database for offline handwriting recognition.
\newblock {\em Int. J. Doc. Anal. Recognit.}, 5:39--46, 2002.

\bibitem{yin2013icdar}
Fei Yin, Qiu-Feng Wang, Xu-Yao Zhang, and Cheng-Lin Liu.
\newblock {ICDAR} 2013 {Chinese} handwriting recognition competition.
\newblock In {\em Proc. ICDAR}, pages 1464--1470. IEEE, 2013.

\bibitem{mouchere2014icfhr}
Harold Mouchere, Christian Viard-Gaudin, Richard Zanibbi, and Utpal Garain.
\newblock {ICFHR} 2014 competition on recognition of on-line handwritten mathematical expressions ({CROHME} 2014).
\newblock In {\em Proc. ICFHR}, pages 791--796. IEEE, 2014.

\bibitem{chi2019complicated}
Zewen Chi, Heyan Huang, Heng-Da Xu, Houjin Yu, Wanxuan Yin, and Xian-Ling Mao.
\newblock Complicated table structure recognition.
\newblock {\em arXiv preprint arXiv:1908.04729}, 2019.

\bibitem{long2021parsing}
Rujiao Long, Wen Wang, Nan Xue, Feiyu Gao, Zhibo Yang, Yongpan Wang, and Gui-Song Xia.
\newblock Parsing table structures in the wild.
\newblock In {\em Proc. ICCV}, pages 944--952, 2021.

\bibitem{jaume2019funsd}
Guillaume Jaume, Hazim Kemal~Ekenel, and Jean-Philippe Thiran.
\newblock {FUNSD}: A dataset for form understanding in noisy scanned documents.
\newblock In {\em Proc. ICDAR {Workshops}}, volume~2, pages 1--6, 2019.

\bibitem{xu2022xfund}
Yiheng Xu, Tengchao Lv, Lei Cui, Guoxin Wang, Yijuan Lu, Dinei Florencio, Cha Zhang, and Furu Wei.
\newblock {XFUND}: A benchmark dataset for multilingual visually rich form understanding.
\newblock In {\em Proc. ACL Findings}, pages 3214--3224, May 2022.

\bibitem{liu2023hidden}
Yuliang Liu, Zhang Li, Hongliang Li, Wenwen Yu, Mingxin Huang, Dezhi Peng, Mingyu Liu, Mingrui Chen, Chunyuan Li, Lianwen Jin, et~al.
\newblock On the hidden mystery of {OCR} in large multimodal models.
\newblock {\em arXiv preprint arXiv:2305.07895}, 2023.

\bibitem{jiang2023revisiting}
Qing Jiang, Jiapeng Wang, Dezhi Peng, Chongyu Liu, and Lianwen Jin.
\newblock Revisiting scene text recognition: A data perspective.
\newblock In {\em Proc. ICCV}, pages 20543--20554, 2023.

\bibitem{yu2023looking}
Wenwen Yu, Mingyu Liu, Biao Yang, Enming Zhang, Deqiang Jiang, Xing Sun, Yuliang Liu, and Xiang Bai.
\newblock Looking and listening: Audio guided text recognition.
\newblock {\em arXiv preprint arXiv:2306.03482}, 2023.

\bibitem{yang2022reading}
Mingkun Yang, Minghui Liao, Pu~Lu, Jing Wang, Shenggao Zhu, Hualin Luo, Qi~Tian, and Xiang Bai.
\newblock Reading and writing: Discriminative and generative modeling for self-supervised text recognition.
\newblock In {\em Proc. ACM MM}, pages 4214--4223, 2022.

\bibitem{sueiras2018offline}
Jorge Sueiras, Victoria Ruiz, Angel Sanchez, and Jose~F Velez.
\newblock Offline continuous handwriting recognition using sequence to sequence neural networks.
\newblock {\em Neurocomputing}, 289:119--128, 2018.

\bibitem{kumari2023comprehensive}
Lalita Kumari, Sukhdeep Singh, Vaibhav Varish~Singh Rathore, and Anuj Sharma.
\newblock A comprehensive handwritten paragraph text recognition system: Lexiconnet.
\newblock In {\em Proc. ICDAR}, pages 226--241. Springer, 2023.

\bibitem{li2023trocr}
Minghao Li, Tengchao Lv, Jingye Chen, Lei Cui, Yijuan Lu, Dinei Florencio, Cha Zhang, Zhoujun Li, and Furu Wei.
\newblock {TrOCR}: Transformer-based optical character recognition with pre-trained models.
\newblock In {\em Proc. AAAI}, volume~37, pages 13094--13102, 2023.

\bibitem{shashank2022improvising}
BN~Shashank, S~Nagesh~Bhattu, and K~Sri Phani~Krishna.
\newblock Improvising the {CNN} feature maps through integration of channel attention for handwritten text recognition.
\newblock In {\em Proc. ICCVAIP}, pages 490--502. Springer, 2022.

\bibitem{wang2020writer}
Zi-Rui Wang, Jun Du, and Jia-Ming Wang.
\newblock Writer-aware {CNN} for parsimonious {HMM-based} offline handwritten {Chinese} text recognition.
\newblock {\em Pattern Recognit.}, 100:107102, 2020.

\bibitem{xie2020high}
Canyu Xie, Songxuan Lai, Qianying Liao, and Lianwen Jin.
\newblock High performance offline handwritten {Chinese} text recognition with a new data preprocessing and augmentation pipeline.
\newblock In {\em Proc. DAS}, pages 45--59. Springer, 2020.

\bibitem{instructblip}
Wenliang Dai, Junnan Li, Dongxu Li, Anthony Meng~Huat Tiong, Junqi Zhao, Weisheng Wang, Boyang Li, Pascale Fung, and Steven Hoi.
\newblock {InstructBLIP}: Towards general-purpose vision-language models with instruction tuning, 2023.

\bibitem{wei2023instructiongpt}
Lai Wei, Zihao Jiang, Weiran Huang, and Lichao Sun.
\newblock {InstructionGPT}-4: A 200-instruction paradigm for fine-tuning {MiniGPT}-4.
\newblock {\em arXiv preprint arXiv:2308.12067}, 2023.

\bibitem{brown2020language}
Tom Brown, Benjamin Mann, Nick Ryder, Melanie Subbiah, Jared~D Kaplan, Prafulla Dhariwal, Arvind Neelakantan, Pranav Shyam, Girish Sastry, Amanda Askell, et~al.
\newblock Language models are few-shot learners.
\newblock {\em Proc. NIPS}, 33:1877--1901, 2020.

\end{thebibliography}

\end{document}